\documentclass[conference]{IEEEtran}

\IEEEoverridecommandlockouts
\usepackage{cite}
\usepackage{amsmath,amssymb,amsfonts}
\usepackage{algorithmic}
\usepackage{graphicx}
\usepackage{textcomp}
\usepackage{xcolor}

\usepackage{svg}

\def\BibTeX{{\rm B\kern-.05em{\sc i\kern-.025em b}\kern-.08em
    T\kern-.1667em\lower.7ex\hbox{E}\kern-.125emX}}
\begin{document}

\title{Spatiotemporal Convolutions on EEG signal \\- A Representation Learning Perspective on Efficient and Explainable EEG Classification with Convolutional Neural Nets}

\author{\IEEEauthorblockN{1\textsuperscript{st} Laurits Dixen}
\IEEEauthorblockA{
\textit{IT University of Copenhagen}\\
Denmark \\
ldix@itu.dk\\
0009-0001-4447-5222}
\and
\IEEEauthorblockN{2\textsuperscript{nd} Stefan Heinrich}
\IEEEauthorblockA{
\textit{IT University of Copenhagen}\\
Denmark \\
stehe@itu.dk\\
0000-0001-9913-3206}
\and
\IEEEauthorblockN{3\textsuperscript{rd} Paolo Burelli}
\IEEEauthorblockA{
\textit{IT University of Copenhagen}\\
Denmark \\
pabu@itu.dk\\
0000-0003-2804-9028}
}

\maketitle

\begin{abstract}
Classification of EEG signals using shallow Convolutional Neural Networks (CNNs) is a prevalent and successful approach across a variety of fields. Most of these models use independent one-dimensional (1D) convolutional layers along the spatial and temporal dimensions, which are concatenated without a non-linear activation layer between. In this paper, we investigate an alternative encoding that operates a bi-dimensional (2D) spatiotemporal convolution. While 2D convolutions are numerically identical to two concatenated 1D convolutions along the two dimensions, the impact on learning is still uncertain. We test 1D and 2D CNNs and a CNN+transformer hybrid model in a low-dimensional (3-channel) and a high-dimensional (22-channel) BCI motor imagery classification task. We observe that 2D convolutions significantly reduce training time in high-dimensional tasks while maintaining performance. We investigate the root of this improvement and find no difference in spectral feature importance. However, a clear pattern emerges in representational similarity across models: 1D and 2D models yield vastly different representational geometries. Overall, we suggest an improved model with a 2D convolutional layer for faster training and inference. We also highlight the importance of architecturally-driven encoding when processing complex multivariate signals, as reflected in internal representations rather than purely in performance metrics.
\end{abstract}


\section{Introduction}

Electroencephalography (EEG) offers a non-invasive and relatively cheap way to record the neural activity of human subjects and has for many decades now been a stable method in cognitive neuroscience. With high temporal resolution, it enables the investigation of the dynamics of human cognition. Still, details of how neural activity is mapped to cognitive processes remain elusive due to the complexity of these patterns. Using deep learning (DL) to aid neural decoding is a promising and active research area, leveraging the strengths of adaptive learning algorithms to effectively classify neural recordings. Convolutional Neural Networks (CNNs) have revolutionised many fields, e.g. computer vision, through their ability to learn complex patterns directly from raw data effectively and efficiently. This success has sparked a growing interest in using CNNs for analysis and interpretation of EEG signals, with potential applications ranging from medical diagnosis \cite{ullahAutomatedSystemEpilepsy2018} to brain-computer interfaces (BCIs) \cite{tabarNovelDeepLearning2016}. However, the unique characteristics of EEG data, such as its dynamic nature, low signal-to-noise ratio, and complex spatiotemporal relationships, present challenges and constraints for traditional CNN design approaches. 

Classification of EEG signals using Artificial Neural Nets (ANN) is an important task in many fields from diagnosis of depression \cite{thoduparambilEEGbasedDeepLearning2020}, autism \cite{radhakrishnanPerformanceAnalysisDeep2021}, dementia \cite{kimDeepLearningbasedEEG2023}, sleep research \cite{phanSleepTransformerAutomaticSleep2022}, stimulus reconstruction \cite{benchetritBrainDecodingRealtime2024}, and brain-computer interfaces (BCI) \cite{zhangSurveyDeepLearning2016}. Existing popular CNN models for EEG analysis employ separate one-dimensional (1D) convolutional layers for handling temporal and spatial dimensions. Notable examples are: FBCSPNet \cite{schirrmeisterDeepLearningConvolutional2017}, EEGNet \cite{lawhernEEGNetCompactConvolutional2018} and the Conformer \cite{songEEGConformerConvolutional2023}. These CNNs, or models of similar design, achieve SOTA performance in EEG classification across a wide array of fields. 

The importance of explainable deep learning in neuroscience cannot be overstated. The ability to interpret the features learnt by deep learning models holds immense potential for advancing our understanding of brain function, and investigating learnt representations from ANNs provides a crucial avenue for future research \cite{kriegeskortePeelingOnionBrain2019}, \cite{vuSharedVisionMachine2018}.

In this article, we investigate the impact on the representational space from replacing the two sequential 1D convolutions with a 2D spatiotemporal convolutional layer that can capture both temporal and spatial information within a single operation. This reduces computational complexity, and therefore training and inference time, something essential when processing multivariate datasets. An analysis of the learnt representations shows that, while learnt spectral features are consistent among model architectures, the underlying representational geometry is significantly changed in 2D models. This has important consequences for researchers looking to use explainability techniques to study learnt feature representation spaces, and underscores the need for investigating architectural inductive biases in encoding layers for EEG decoder models.

\section{Background}

The strength of ANN models lies in their ability to perform feature extraction and classification in an end-to-end fashion. This removes the need for manual feature extraction, which can be time-consuming and risks introducing human biases. It is therefore desired to design ANNs that can work on raw data for classification, with as little preprocessing or feature extraction as possible before training. Different ANN architectures provide different structure-encoded constraints, sometimes referred to as inductive biases \cite{battagliaRelationalInductiveBiases2018}. It is therefore paramount to design ANNs with the data structure in mind, so that they capture the relevant dimensions and features well. With this motivation, we take a closer look at the data structure of raw EEG signals and analyse how convolutions are used on them. 

Raw EEG recordings are characterised by both a temporal and a spatial dimension. EEG models are often made with these dimensions as the guiding idea. The spatial dimension comes from the physical EEG sensor electrodes being placed in specified positions on the scalp of subjects, typically following a standardised montage scheme, like the popular 10-20 system, securing consistent electrode positions across different subjects and sessions. Each electrode records a continuous streaming signal, measuring the electrical potential on the scalp relative to a specified reference electrode. These data streams are then recorded and denoted as EEG channels. Sequences of raw EEG signals are then characterised as a multivariate time series $\mathcal{X} = \mathbb{R}^{C\times T}$ with $C$ as the number of channels and $T$ time points in a recording. 
In a classification task, each sample is accompanied by labels $y\in\mathcal{Y}$, where $\mathcal{Y}$ is dependent on the given task. For example, in a binary motor imagery (MI) task, the labels might be $\mathcal{Y} = [\text{"left hand"}, \text{"right hand"}]$. The goal for a discriminative model $\mathcal{M}$ is to approximate a function $f: \mathcal{X} \rightarrow \mathcal{Y}$ that minimises the distance from predicted labels $\hat{y}$ to true labels $y$. Since EEG sampling frequency typically is high ($\geq 250$ Hz), $T$ tends to be much larger than $C$, making any sample $x\in \mathcal{X}$ imbalanced in dimensions. Models training on $\mathcal{X}$ is then challenged with finding informative features along both a temporal and a spatial orientation. EEG signals have a low signal-to-noise ratio, so this feature-extraction step is especially important here. Some methods sidestep the temporal dimension of this by doing feature extraction as a preprocessing step by transforming the data into frequency bands, typically the common five, which are known to hold relevant information about brain states \cite{chaddadElectroencephalographySignalProcessing2023}. Introducing spectral features this way removes the complexity of handling time series, but adds the potential bias of relying on specific bands and performing a heavy dimensionality reduction. In this paper, we focus on raw EEG signals and will proceed accordingly.

\subsection{Models of space and time}
The most common way to solve this two-dimensional problem is to create separate modules of the model that handle each dimension, namely a temporal module and a spatial module. These 'modules' can be single- or multi-layer, and their positions are, in theory, arbitrary. These modules can be, and often are, followed by fully connected or other types of layers, but these two modules are necessary to include at some stage in the model. In the DL literature, common module types for handling the temporal dimension in sequential or time-series data include LSTM \cite{alhagryEmotionRecognitionBased2017}, RNN  \cite{zhangDeepLearningEEGBased2021}, transformer \cite{phanSleepTransformerAutomaticSleep2022}, and the 1D convolutional layer \cite{zhouEpilepticSeizureDetection2018}. Common layer architectures for the spatial module are graph convolutional layers \cite{songEEGEmotionRecognition2020}, spatial attention layers \cite{benchetritBrainDecodingRealtime2024}, and also 1D convolutional layers. Here, a common and successful model for EEG signals is a pair of 1D convolutional architectures. The most popular being the EEGNet \cite{lawhernEEGNetCompactConvolutional2018} and FBSCPshallowNet~\cite{schirrmeisterDeepLearningConvolutional2017}. Both are characterised by their few layers and temporal + spatial 1D convolutional layers, sometimes followed by more convolutions, and sometimes by a fully connected classification layer. It is common to mix and match different components to create new and hopefully better models, see e.g. \cite{liCrossSubjectEEGEmotion2021} using 1D convolution along the temporal dimension and a graph structure along the spatial. 

\subsection{Convolutions on EEG signal}
Since convolutional layers are very efficient in number of free parameters, these models retain a fast training speed, relatively low data requirements and strong performance on many different tasks. Due to this, they are often easy to use and have gained large popularity. Thus, the focus of this paper is specifically on these convolutional layers and in the following, we will lay out in detail how they work in encoding raw EEG signals. They are referred to as 1D convolutional layers here since they only convolve over a single dimension, i.e. the temporal or the spatial. Temporal convolutional kernels are defined by $$k_{t}\in \mathbb{R}^{1\times m}$$ with $m$ as the kernel size. Then the convolutional operation on data point $x \in \mathbb{R}^{C\times T}$ is defined as $$h_{t}(t)=(x*k_{t})(t) = \sum^m_{\tau=-m}x(t-\tau)\cdot k_t(\tau)$$ To create a spatial 1D convolution across channels we fix the kernel size to 1 in the temporal dimension, but define the kernel size along the spatial dimension as the number of channels: $$k_s \in\mathbb{R}^{C \times 1 }$$ and the convolution becomes: 
$$h_s(t)=(x*k_s)(t) = x(t)\cdot k_s$$
which is simply the dot product of the kernel and signal values of all channels at time $t$, or equivalently, a weighted sum of the channels.

It is worth noting here the reason why we are not using a kernel size smaller than $C$. Convolutional layers rely on sparse interactions along the dimension they are convolving, utilising the fact that local signals are related, resulting in efficiency through parameter sharing \cite{Goodfellow-et-al-2016}. This is very useful along the temporal dimension, as mentioned above, $T$ is often quite large in EEG signals. However, this locality assumption does not hold for channels, as we cannot meaningfully order the EEG electrodes along a single dimension. This is not a problem, however, as the number of channels is usually not very large ($\leq 144$), so we do not suffer from poor scaling. We now investigate the 2D spatiotemporal convolutional operation for EEG signals. This is simply a combination of the two 1D convolutions introduced above, we write $$k_{st}\in\mathbb{R}^{C \times m}$$and the convolutional operation as $$h_{st}(t)=(x*k_{st})(t) = \sum^m_{\tau=-m}x(t-\tau)\cdot k_{st}(\tau)$$which closely resembles the definition of $h_t(t)$, with the extension the size of the kernel, adding weighted sum of the channels in the same operation. This also means that $$h_s(h_t(t)) = h_{st}(t)$$provided identical kernel weights, and we thereby preserve the shape of our data after applying the transforms. Are we losing representational power by adding this collapse? Only in situations where we apply operations between $h_s$ and $h_t$, which might be the case for some model architectures. But in cases where no activation function is applied in between, these two operations should exhibit identical representational power. This is indeed the case in the aforementioned popular models, EEGnet and FBSCPshallowNet. Of additional note is the EEG Conformer model, a very popular type of model type introduced by Song et al, 2023 \cite{songEEGConformerConvolutional2023}, designed to apply the strength of the transformer architecture to EEG signals. The idea is to first transform the input data through convolutional layers and reduce it to a shape conducive to the transformer. The main strength of the Conformer compared to CNNs is the ability to encode longer-ranged temporal relations, as convolution must rely on pooling to summarise across the temporal dimension. Here, again, the convolutional layers are structured like the FBSCPshallowNet, without activation or other operations between the temporal and spatial convolutions. 

\subsection{Representation learning} 
Representation learning is a machine learning technique in which a model learns to automatically discover meaningful patterns and features within a given dataset. Instead of relying on handcrafted features, the model learns to extract these representations directly from the raw data. This approach is particularly powerful for complex data such as electroencephalography (EEG) signals, which are often noisy, high-dimensional, and exhibit complex temporal and spatial relationships that are difficult to interpret manually. Representation learning helps models to identify informative, lower-dimensional features that retain essential information while filtering out noise, thus making the learning process more efficient and the results more comprehensible. Moreover, investigating the pattern automatically extracted via representation learning can help explain the decision-making process of the overall model. Explainability is increasingly important in machine learning as it is applied in new fields; unfortunately, as modern models grow larger and more effective, their explainability naturally decreases. Today, many treat artificial neural networks (ANN) as black box models, yielding little to no insight about the underlying system. This can, however, be combated by employing techniques from the explainable AI (XAI) literature. In the application of machine learning to neuroscience and particularly EEG, explainability plays a crucial role in the ability of any model to give insights into the inner functioning of the brain. Therefore, in this paper, we take a representation-learning stance and highlight not only performance but also the learnt representations as an integral part of model building and evaluation. The goal is to create strong models of brain signals that also provide insight into the underlying neural activity and cognitive states. 

\section{Model}

To assess the impact of the proposed spatiotemporal filter, we have modified two state-of-the-art models (a shallow CNN based on the ShallowFBCSPNet~\cite{schirrmeisterDeepLearningConvolutional2017}, and a more complex conformer model inspired by the EEG Conformer~\cite{songEEGConformerConvolutional2023}) by changing their initial convolutional layer, and we compare their performance against their base version. This section introduces the specific architectures of these models and explains the rationale for the authors' decision to explore differences between 1D and 2D convolutional approaches.

\subsection{ShallowFBCSPNet}

The convolutional neural networks (CNN) used here are based on the ShallowFBCSPNet \cite{schirrmeisterDeepLearningConvolutional2017}, a simple shallow CNN with three main layers. One layer of temporal convolution on a single channel, one layer of convolution using all channels and a fully connected classification layer. 40 kernels were used with a kernel size of 25 along the time dimension. The Elu nonlinear activation function was used after the second convolutional layer; it is important to note that it is not used after the first convolutional layer. A pooling layer is added after the activation function with a pooling size of 100. Then a batch normalisation layer, and a dropout layer before the fully connected layer. 

\subsubsection{Conformer}
The conformer model used here is based on the original EEG Conformer \cite{songEEGEmotionRecognition2020}. It consists of a block of early convolutional layers with a transformer added before classification. The early convolutional layers are exactly the same structure as the CNN described above, up to the fully connected layer. A 2-headed transformer architecture is used here, following two fully connected layers leading to the output layer. Conformers are significantly bigger than the CNNs but should offer better performance on larger datasets. 

\subsubsection{2D models}
The 2D models are the main independent variable in this paper, as we aim to isolate the effect of adding spatiotemporal learning rather than splitting them into spatial and temporal layers. The 2D models are referred to as such since they combine the two 1D convolutional layers of the previously described models into a single layer. Illustrated in the figure, we see the difference between 1D and 2D convolutions. Important to note, though, is that in neither case does the convolutional operation slide across channels. Instead, in the 1D spatial and 2D temporospatial layers, all EEG channels are included in the filter region of interest as it 'slides' across the time dimension. 

\begin{figure}
    \centering
    \includegraphics[width=1\linewidth]{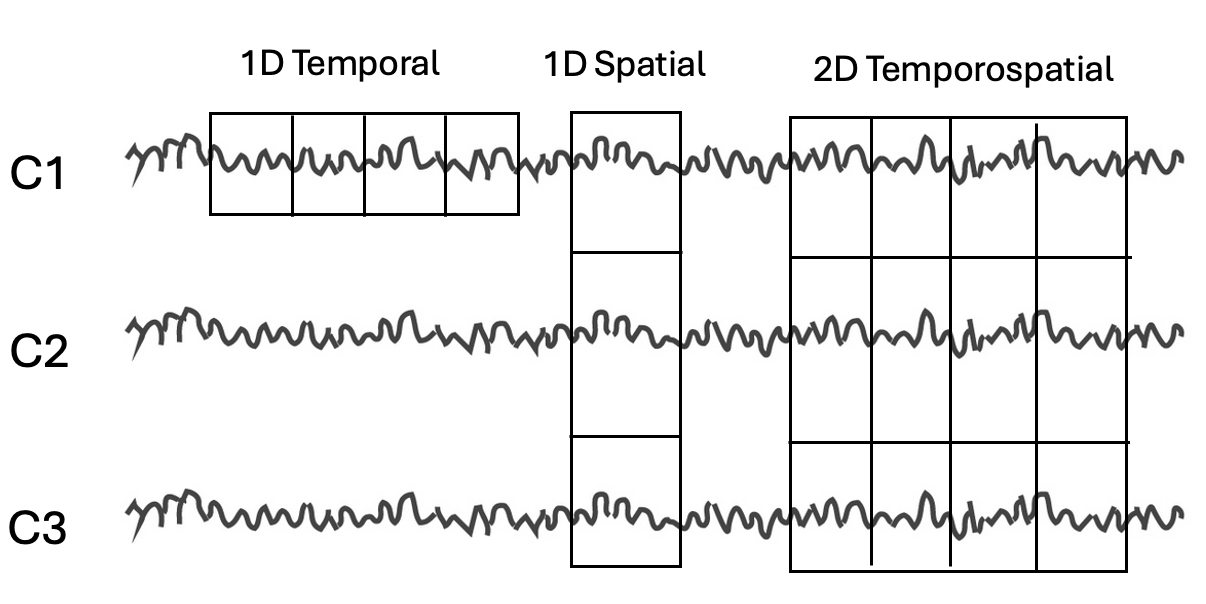}
    \caption{1D and 2D convolution illustration}
    \label{fig:Pasted image 20241106100948.png}
\end{figure}

\section{Evaluation}

The performance and interpretability of 1D and 2D convolutional architectures for EEG decoding are evaluated using accuracy, training time, feature representation analysis, and representational similarity analysis (RSA). The models are trained and compared on a dataset designed for a motor-imagery BCI task.

\subsection{Datasets and tasks}
The datasets used for this testing are taken from the BCI benchmarking framework MOABB~\cite{jayaramMOABBTrustworthyAlgorithm2018}. MOABB provides standardised processing of various datasets, enabling rigorous model testing and making it convenient. Two datasets were chosen for this paper, both from the BNCI Competition IV \cite{tangermannReviewBCICompetition2012}. The two datasets (BNCI2014-001 and BNCI2014-004 on MOABB) have been set up in the BCI scheme but differ slightly in parameters, such as the number of classes and channels, which are important for understanding model performance across different types of datasets. Both datasets use the BCI framework for motor imagery, in which nine subjects imagine a movement after a cue is shown. They maintained this imagining for 4-4.5 seconds and, in some trials, received their classification performance. BNCI2014-001 \cite{tangermannReviewBCICompetition2012}, referred to as dataset 1 from now, is the larger and harder of the two tasks. It has four classes, 22 channels and 62208 total trials. BNCI2014-004, referred to as dataset 2 from now, is smaller but easier. It has 3 channels, but is a binary classification task with 32400 total trials. These two datasets offer slightly different settings but are as comparable as possible, allowing us to compare and contrast model performance across them. The task for both datasets is a mixed-subject generalisation task. That is, we combine all subjects' trials so that each subject is present in both training and testing. The preprocessing was kept as MOABB's standard processing for these datasets, a bandpass filter between 8 Hz and 32 Hz was applied along with a standard scaler, and no baselining was applied. All models are evaluated using cross-entropy loss and accuracy in a multi-class classification setting. 

\subsection{Training procedure and hardware}
We trained all models using a 5-fold cross-validation scheme, with 20\% of the total trials in the test set and stratified by label to maintain class balance. Crucially, this produces five versions of each model, each serving as a repetition of that model type, allowing us to compare them. The models are trained under the same hyperparameters, learning rate of 0.0001, batch size of 128, using the ADAM optimiser. The training utilised an early stopping scheme on a 20\% validation split from the training data. If 100 epochs passed without improvement on the validation loss, the training would stop, and the best model would be saved. A hard cap of 3000 epochs was added, but was never reached during any training. For regularisation, a weight decay of 0.0001 was used, along with a dropout rate of 0.5 during training. All models were trained on an NVIDIA V100 graphics processing unit with 32 GB of memory, which was sufficient to hold the full dataset and models, so no time was spent loading data from disk. 

\subsection{Extraction of signal features and kernel activations}
Within a representation-learning framework, we want to understand which features the models have learnt. Therefore, we extract features from each trial in the dataset, then run those trials through the models and record the model activations. We can then compare these features and activations to understand the model's representations. The features extracted here are band-power features, which are widely used in EEG analysis and are often also used as non-temporal features in deep learning models. Since the data is filtered for alpha and beta bands (8-32 Hz) in preprocessing, we divide the data into 5 bands: [8-10, 10-12, 12-16, 16-24, 24-32] all in Hz, and extract them using MNE's \cite{gramfortMEGEEGData2013} implementation of calculating the power spectrum density using Welch's method of the Fast Fourier Transform \cite{cochranWhatFastFourier1967}. 
The features are extracted from the full trial, so no temporal dynamics are at play. The activations of both CNN, conformers and their 2D versions are extracted at the same point into the model, to ensure as much comparability as possible. The activations are taken after the pooling layer, that is, after the convolutional and activation layers, and before the fully connected or transformer layers. We have also turned off batch normalisation and dropout during representation recording, as these are meant for training and would disrupt the analysis. At this point, all models' output dimensions are identical and can be compared directly. 

\subsection{Representational similarity analysis}
Representational similarity analysis (RSA) \cite{kriegeskorteRepresentationalSimilarityAnalysis2008} is a popular technique in computational neuroscience for comparing representations across different recording modalities. Similar methods have also been used to compare models across different weight initialisations and architectures in the DL literature \cite{kornblithSimilarityNeuralNetwork2019}. We calculate the average distance of model activation patterns to the same data points. Here, we simply use all samples of each dataset. All model types and folds are compared and combined into a representational dissimilarity matrix (RDM), which can be visualised and inspected to reveal differences in representations between model types. Activations are measured after the pooling layer of each model and flattened for easy comparison. 

\section{Results}

\begin{figure}
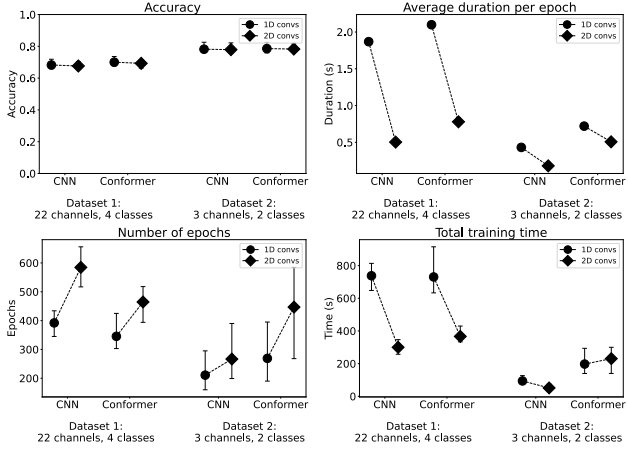

    \centering
    \includesvg[width=0.46\linewidth]{imgs/performance/accuracy.svg}
    \includesvg[width=0.46\linewidth]{imgs/performance/dur_avg.svg}
    \includesvg[width=0.46\linewidth]{imgs/performance/num_epochs.svg}
    \includesvg[width=0.46\linewidth]{imgs/performance/run_time.svg}
    \caption{Performance results of 1D (circles) and 2D (squares) models are compared across two datasets.}
    \label{fig:performance_results}
\end{figure}

In Figure \ref{fig:performance_results}, we compare the 1D and 2D versions of both the shallow CNN and the EEG Conformer model on the two datasets. We observe two main results. First, the model's accuracy is unchanged when comparing 1D to 2D models. We also have very little variance in accuracy across folds for the same model, indicating the reliability of the results. Additionally, we see an increased accuracy on dataset 2; this is likely due to the easier task of binary classification compared to four-way classification. The second observation is that 2D models are significantly faster to train on dataset 1 and that they are similarly or slightly faster on dataset 2. We see that this increase in faster training is due to the time per epoch being much lower for 2D models, especially on dataset 1. We also see that the number of epochs required to reach the training limit (see the section on the training procedure for details) is higher for the 2D model. However, as shown in the total training time, the 2D models are much faster to train on dataset 1. Why do we observe the difference between datasets 1 and 2? The difference very likely stems from the difference in the number of channels. Dataset 1 has 22 EEG channels compared to Dataset 2's 3 EEG channels, making the input data dimensions quite different. We will further discuss the differences in later sections. 

\subsection{Feature reconstruction}

\begin{figure}
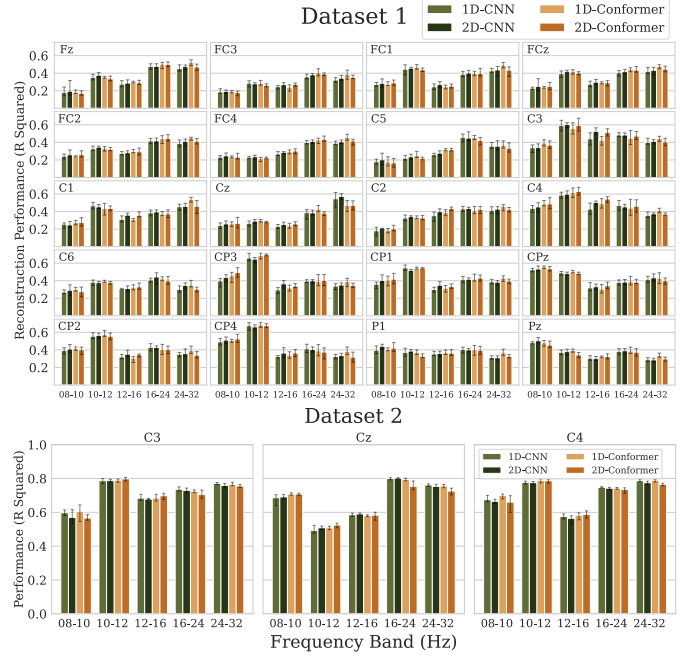

    \centering
    \includesvg[width=1\linewidth]{imgs/band_power_reconstruction-BNCI_001.svg}
    \includesvg[width=1\linewidth]{imgs/band_power_reconstruction_BNCI-004.svg}
    \caption{Feature reconstruction results. While significant differences can be seen between reconstruction success between channels and bands, no differences is observed between model types.}
    \label{fig:reconstruction.svg}
\end{figure}

To assess whether models have learnt different features of the dataset, we reconstruct band-power features in subdivisions of the alpha and beta bands from the activation signals after the 1D or 2D convolutional layers using a linear regression model. Figure \ref{fig:reconstruction.svg} shows the results of the band power reconstruction, containing channel-by-channel R-squared values for each band. The first observation is the ability to reconstruct band powers with varying success from kernel activations across different channels. This shows that the models have, unsurprisingly, learnt spectral features of the data during their training. We also see that the models have to different degrees learnt different features from different channels. For example, in dataset 1, the R-squared value for C3 in frequency band 10-12 Hz is around $0.8$, whereas in the band 8-10 Hz we see $<0.6$ R-squared. An opposite pattern is then observed in channel Cz, where the R-squared values are higher in 8-10 Hz than in the 10-12 Hz. One could study further and analyse these differences to understand which frequency ranges in which channels are important for motor imagery classification. However, we are interested here in observing differences between model types, especially between 1D and 2D. We observe no significant differences across almost all channels and frequency bands. Certainly, no systematic differences can be observed. This leads us to the conclusion that the models have essentially learnt and retained the same information about band powers in their convolutional layers. Lastly, we also observe how R-squared values are generally lower in dataset 2. This is likely due to the many more channels in dataset 2, so any particular feature in a given channel out of 22 will contribute less to the overall activation than in dataset 1, where only three channels are present. 

\subsection{Correlational feature analysis}

\begin{figure}
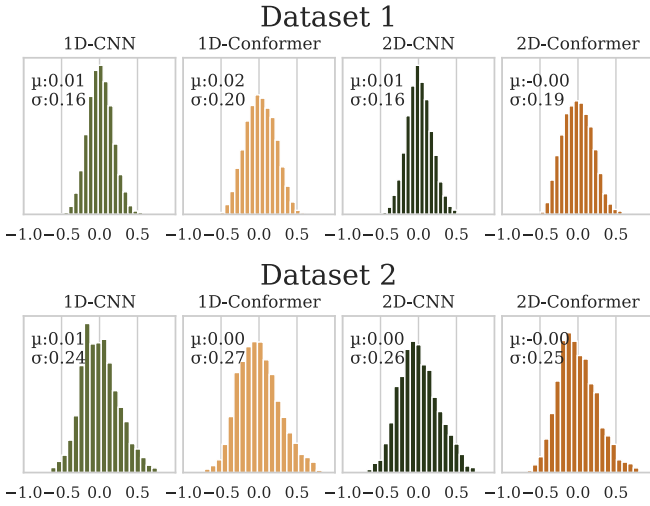

    \centering
    \includesvg[width=1\linewidth]{imgs/correlation_average_d1.svg}
    \includesvg[width=1\linewidth]{imgs/correlation_average_d2.svg}
    \caption{Distribution of correlations}
    \label{fig:correlation_average.svg}
\end{figure}

To better understand how this information is represented in the activations, we perform a supplementary correlational analysis. Figure \ref{fig:correlation_average.svg} shows the histograms of individual kernel-band power correlation per model on the two datasets. All folds and bands are mixed in this plot to simplify reading, as no patterns of note were observed between them. Instead, we see a similar pattern as shown in the reconstruction analysis. Dataset 2 shows clearly higher correlations between individual kernel activations and spectral features than Dataset 1. This can be seen by the fact that while the mean ($\mu$) of all of these distributions is very near 0, the standard deviation ($\sigma$) is higher in dataset 2 (0.24, 0.27, 0.26, 0.25) than in dataset 1 (0.16, 0.2, 0.16, 0.19) for the four model types, respectively. As argued above, this seems likely to stem from the fact that signal features from only three channels are distributed across 40 kernels, resulting in lower overall correlations. These results underline the necessity of reconstructing analysis across all kernels, as even though the feature information is spread across dataset 1, the collective activation among kernels is sufficient to reconstruct the features. 

\subsection{Representational similarity analysis}

\begin{figure}
    \centering
    \includesvg[width=0.99\linewidth]{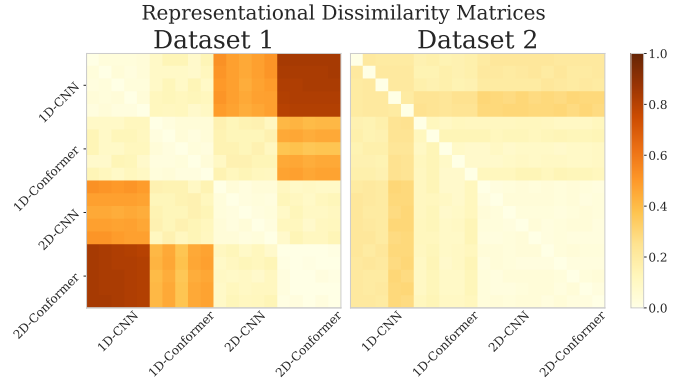}
    \caption{Representational Dissimilarity Matrices (RDMs) of datasets 1 and 2. Five folds of each model type representational dissimilarity is shown. Higher values indicate dissimilarity.}
    \label{fig:rdms}
\end{figure}

The RSA performed on the two datasets compares the similarity of the models' activations after the convolutional layers. We examine the 5 folds of each model to assess how stable the model architectures are relative to each other and across folds of the same model. The RDMs (fig. \ref{fig:rdms})of the two datasets show different overall patterns. The first observation is that in almost all cases, models are more similar to different folds of the same type than to folds of other models. This is true for all comparisons except the 1D models in dataset 2. We also see that the change of convolutional dimension from 1D to 2D explains some similarities. This is best illustrated in the 2D models of dataset 2, which are very similar. In dataset 1, the patterns are very clear, CNNs are closer to each other, and 1D models are similar to 2D models. Another important observation is that dissimilarity is overall much lower in dataset 2. The dissimilarity between 1D models and 2D models is much higher in Dataset 1 than in Dataset 2. This is likely explained by the same arguments highlighted in the previous sections. Dataset 1 has 22 channels, and Dataset 2 has 3. This means individual activations are overall smaller in Dataset 2, as seen in the correlational analysis (fig. \ref{fig:correlation_average.svg}). The RSA shows that when there are a large number of channels, the representational similarity is quite different between 1D and 2D models. This is somewhat surprising, since the performance, reconstruction, and correlational analyses seemed to show little difference between the model types. This suggests that while the same features are learnt from the dataset and performance is similar, the representation in a 2D setting is quite different, supporting the idea that 2D layers are not simply equivalent to two 1D layers.

\section{Discussion}

\subsection{Representational dissimilarity}
We performed multiple analyses of the learnt representations of the different model architectures. One not-so-surprising finding is that the learnt spectral features of the 1D and 2D convolutional layers were practically identical. The reconstruction analysis showed that while models learnt different features from different channels, there were no differences between 1D and 2D models. Since we argued that the representational power should be the same in the 2D convolutional layer as in two 1D layers, this result is in line with our hypotheses. More surprising is the stark difference in the representational dissimilarity in dataset 1. Here, we see low dissimilarity between folds of the same model but large differences between model types. With both CNN-to-conformer and 1D-to-2D changes making a big difference, the highest dissimilarity was observed between the 1D CNN and the 2D conformer. This shows that while the models have learnt the same underlying features of the data, as measured by frequency bands, the representations of these features differ significantly across model types. This difference was not as pronounced in dataset 2, but as already discussed, some of this is likely due to the low-dimensional signal ($C=3$) being spread out on a higher number of kernels ($K=40$), resulting in less visible effects. Regardless, the results suggest that, under complex multivariate decoding tasks, representational geometry is heavily influenced by inductive biases in the encoding architecture. 
This large difference in representational similarity shows that models have found other ways to encode the same information in their weights, leading to altered training procedures and learning capabilities. 

\subsection{Training Speed Increase}
The findings from the model evaluation showed that 2D convolutional models performed identically well on both datasets in terms of accuracy, but their total training time was significantly lower in the dataset with 22 channels and slightly lower in the dataset with 3 channels. This reduction in training time comes from a much faster pass through the model, as shown by the per-epoch duration. The effect was mitigated by requiring more epochs to achieve stable performance on the validation set. This provides evidence that the 2D convolutional layer is a more efficient early representation layer for EEG signals with a large number of channels. Or more likely, the effect is still present but weaker in datasets with low channel numbers. An easy answer to why epochs were faster with the 2D operation is that it replaced two separate sequential operations. Training speed is sometimes an undervalued factor in the machine learning literature. It is often the limiting factor in many practical applications, rivalled only by access to large data volumes. Often, the remedy for slow training is just buying access to high-quality GPUs, but that is not a viable strategy for all research facilities or companies with limited resources. Of central importance is also the environmental impact of training large DL models, and efficient training is a clear way to reduce the total energy expended during development and inference, if one is interested in applying these methods at scale. Training time also plays a crucial role in the development and evaluation of models. Poor generalisability of results may stem from limited testing across RNG seeds or other datasets and tasks. Significantly reducing the training time of both simple CNN models and the more complex Conformer model enables higher generalisability and more robust results in the research community. The last important application of this finding is not directly training time but inference time, where we also found the most drastic improvement with 2D convolutions. Slow inference is especially crippling in BCI, where online evaluation and fast feedback are essential. Especially with a higher number of channels and consumer-grade computing power, speed can be a practical barrier for a usable BCI application. The results from this work suggest that using 2D convolutional layers in both shallow CNNs and conformers would be a much more computationally efficient choice going forward.

Some limitations of this work are worth highlighting. We only tested the models on two datasets with very similar task settings. Further work should explore generalisability to other task settings. Another set of tests not included here, primarily for clarity's sake, is testing with different model sizes, including adding more layers and changing the number of kernels, batch size, and learning rates. A systematic test across all different hyperparameters might make results much less clear to communicate, but should be considered going forward, when applying these ideas to new research. We chose to use the models' default settings as a baseline and first test to make the results as relevant to other practitioners as possible. 

\section{Conclusions}

In this work, we conducted a series of empirical tests using two of the most popular model architectures to classify EEG signals and performed an analysis of the learnt representations. We demonstrate that using a 2D spatiotemporal convolution in place of two separate 1D convolutions can significantly reduce training time for both Shallow CNNs and Conformer models while maintaining identical performance. Faster training times can facilitate more extensive experimentation with different model architectures and hyperparameters, potentially leading to the development of more robust and generalisable models.

This work demonstrates that switching from sequential 1D convolutional layers to unified 2D spatiotemporal convolutional layers in EEG neural network models has a profound impact on the geometry of internal representations. While both architectures learn comparable spectral features, our analyses reveal that 2D convolutions restructure representational space in distinctly different ways, especially as data dimensionality grows, offering new avenues for explainability and interpretation in neural signal processing beyond mere performance gains.

\bibliographystyle{IEEEtran}
\bibliography{references}

\end{document}